\documentclass[runningheads]{llncs}
\usepackage[T1]{fontenc}
\usepackage{amsmath,amssymb,amsfonts}
\usepackage{algorithmic}
\usepackage{graphicx}
\usepackage{textcomp}
\usepackage{graphicx}
\usepackage{booktabs}
\usepackage{multirow}
\usepackage{graphicx,verbatim}
\usepackage{makecell}
\usepackage{multirow}
\usepackage{xcolor}
\usepackage{hyperref}
\begin{document}

\title{Bipartite Patient-Modality Graph Learning with Event-Conditional Modelling of Censoring for Cancer Survival Prediction}

\author{Hailin Yue\inst{1}\index{Yue, Hailin} \and Hulin Kuang\inst{1}\thanks{Corresponding author}\index{Kuang, Hulin}\and Jin Liu\inst{1,2}\index{Liu, Jin}\and Junjian Li\inst{1}\index{Li, Junjian}\and Lanlan Wang\inst{1}\index{Wang, Lanlan}\and Mengshen He\inst{1}\index{He Mengshen}\and Jianxin Wang\inst{1,2}\index{Wang, Jianxin}}  
\authorrunning{H. Yue et al.}
\institute{\textsuperscript{1}Hunan Provincial Key Lab on Bioinformatics, School of Computer Science and Engineering, Central South University, Changsha, China \\ 
\textsuperscript{2}Xinjiang Engineering Research Center of Big Data and Intelligent Software, School of Software, Xinjiang University, Urumqi, China \\
 \email{hulinkuang@csu.edu.cn}}

\maketitle              
\begin{abstract}
Accurately predicting the survival of cancer patients is crucial for personalized treatment. However, existing studies focus solely on the relationships between samples with known survival risks, without fully leveraging the value of censored samples. Furthermore, these studies may suffer performance degradation in modality-missing scenarios and even struggle during the inference process. In this study, we propose a bipartite patient-modality graph learning with event-conditional modelling of censoring for cancer survival prediction (CenSurv). Specifically, we first use graph structure to model multimodal data and obtain representation. Then, to alleviate performance degradation in modality-missing scenarios, we design a bipartite graph to simulate the patient-modality relationship in various modality-missing scenarios and leverage a complete-incomplete alignment strategy to explore modality-agnostic features. Finally, we design a plug-and-play event-conditional modeling of censoring (ECMC) that selects reliable censored data using dynamic momentum accumulation confidences, assigns more accurate survival times to these censored data, and incorporates them as uncensored data into training. Comprehensive evaluations on 5 publicly cancer datasets showcase the superiority of CenSurv over the best state-of-the-art by 3.1\% in terms of the mean C-index, while also exhibiting excellent robustness under various modality-missing scenarios. In addition, using the plug-and-play ECMC module, the mean C-index of 8 baselines increased by 1.3\% across 5 datasets. Code of CenSurv is available at https://github.com/yuehailin/CenSurv.


\end{abstract}
\section{Introduction}

Globally, cancer is a huge public health challenge. According to data from the World Health Organization (WHO), millions of people are diagnosed with cancer each year, leading to millions of deaths~\cite{sung2021global}. Personalized treatment for patients is recommended in clinical practice~\cite{psutka2022clinical,lee2024machine}. To guide personalized treatment for patients, it is necessary to accurately predict the survival of cancer patients.

\begin{figure}[ht!]
\centerline{\includegraphics[width=0.8\textwidth]{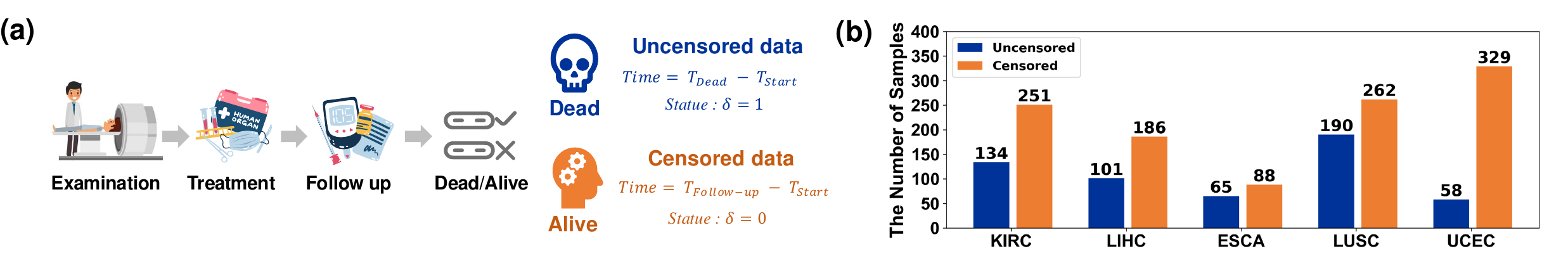}}
\caption{(a) The distinction of censored and uncensored data. For uncensored data, the survival status is dead ($\delta$ = 0), and the survival time is accurate. For censored data, the survival status is alive ($\delta$=1), and the survival time calculated with the follow-up time as the event endpoint is inaccurate and smaller than the actual survival time. (b) The number of censored and uncensored data on 5 datasets in our study.}
\label{six_data}   
\end{figure}

With advances in medical imaging and genomics, many studies have identified potential markers for cancer survival prediction~\cite{shao2023hvtsurv,elforaici2025semi}. For example, Shao et al. proposed a hierarchical vision transformer for survival prediction using whole slide images~\cite{shao2023hvtsurv}. Increasingly, studies combined imaging, genetic, and clinical data for more accurate survival prediction~\cite{hou2023hybrid,liu2024agnostic}. For example, Chen et al. used both pathological images and genetic data for survival prediction~\cite{chen2020pathomic}. Other methods like MultiSurv~\cite{vale2021long}, SurvMamba~\cite{chen2024survmamba} and SurvPath~\cite{jaume2023modeling} also used these multimodal data for survival prediction. However, these methods cannot adapt to the modality-missing scenarios. The existing methods for missing modalities overlook the differences between complete and incomplete modalities~\cite{hou2023hybrid,qu2024multi}. Thus, a method that explores differences between complete and incomplete modalities while remaining robust in modality-missing scenarios is essential.

In clinical practice, the number of uncensored samples with accurate survival times is limited, with the majority consisting of censored data lacking precise survival times (the specific details are shown in Fig. \ref{six_data}). Nowadays, most studies on survival prediction only consider the relationship between uncensored samples and other samples with known survival risks, while ignoring the potential value of censored samples with uncertain survival risks~\cite{jaume2023modeling,shao2023fam3l}. Existing studies using censored data do not consider the impact of the true survival time of censored samples on model performance~\cite{shahin2024centime}. Therefore, effectively mining the actual survival time of unavoidable censored data, updating the corresponding survival status, and incorporating these data into training as uncensored to increase the proportion of uncensored samples may enhance the performance of models in survival prediction.

Based on these considerations, we propose a novel CenSurv for cancer survival prediction. CenSurv proposes a bipartite graph to simulate the patient-modality relationship and explore modality-agnostic features. In addition, CenSurv introduces event-conditional modelling of censoring (ECMC) to fully use of the censored data. The main contributions of CenSurv can be summarized as follows: 1) We propose a plug-and-play ECMC to uncover the value of censored data, improving the mean C-index by 1.3\% across 8 baselines evaluated on 5 cancer datasets. 2) We propose a bipartite patient-modality graph that fully exploits the flexibility of graph learning and enables the model to not only work with different modality-missing scenarios but also achieve better performance than existing methods for handling modality-missing.

\begin{figure*}[t!]
\centerline{\includegraphics[width=1.0\textwidth]{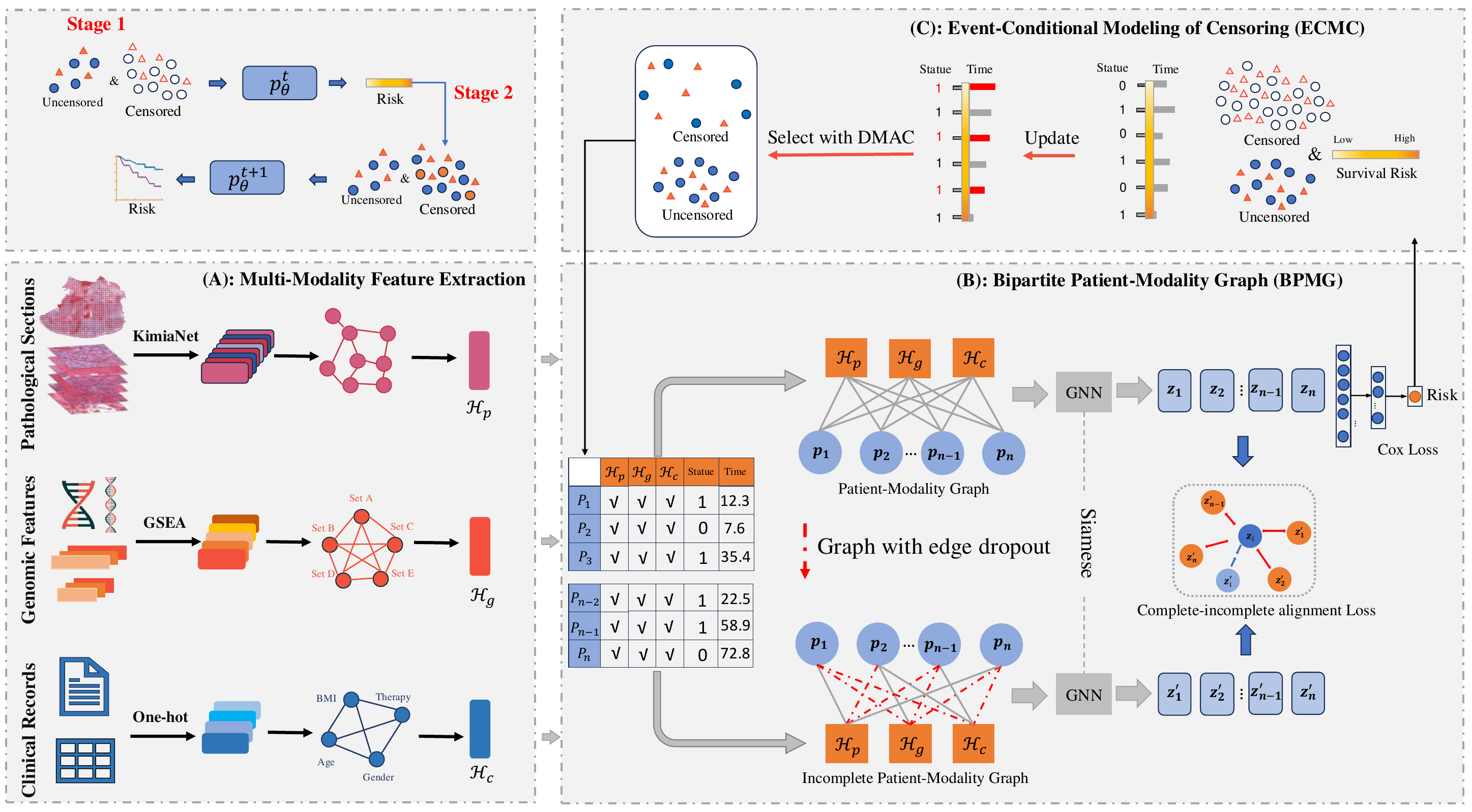}}
\caption{Overview of the bipartite patient-modality graph learning with event-conditional modelling of censoring (CenSurv).}
\label{fig2}
\end{figure*}

\section{Methods}

The goal of this study is to use pathological sections, genomic features, and clinical records for survival prediction. Survival prediction, also known as time-to-event analysis, uses both survival time and survival status to assess the probability of experiencing a specified event, such as mortality in the clinical setting before a time. An overview of CenSurv is shown in Fig. \ref{fig2}. CenSurv uses the bipartite patient-modality graph with event-conditional modelling of censoring to improve the performance of the model in survival prediction and its robustness under different modality-missing scenarios. CenSurv consists of multimodal feature extraction (Fig. \ref{fig2} (A)), bipartite patient-modality graph (Fig. \ref{fig2} (B)), and event-conditional modelling of censoring (Fig. \ref{fig2} (C)).

\subsection{Multimodal Feature Extraction}

 For the pathological section $\operatorname{(p)}$, we crop the original images at 10× magnification into non-overlapping 512$\times$512 patches. We then use the pre-trained KimiaNet~\cite{riasatian2021fine} to extract features for each patch. Since survival prediction requires both local and global tumor features~\cite{hou2023hybrid}, we use a graph to represent both the local and global information of pathological sections. Specifically, each patch is connected to its 8 spatially adjacent patches, forming a graph of the entire pathological section. For the genomic profile $\operatorname{(g)}$, we generate 5 genomic embeddings using Gene Set Enrichment Analysis (GSEA)~\cite{subramanian2005gene}. These embeddings are then connected to form a graph representation of the genomic profile. For clinical records $\operatorname{(c)}$, we one-hot encode each clinical feature to obtain a vector representation, which we fully concatenate to create a graph representation of the clinical data. To ensure consistency across modalities, we align all node features to 1024 dimensions via zero padding. Subsequently, we use three independent GraphSAGE~\cite{hamilton2017inductive} models to facilitate information transfer across each modality graph.

To unify the dimensions of different modalities, we introduce hyperedge. Specifically, we first link all nodes belonging to the same modality by a hyperedge. Then, we use the attention pooling operation to extract the high-order representation of each modality. The representation of modality $m$ with $N$ nodes can be implemented as:

\begin{equation}
\mathcal{H}_m=\sum_{n=1}^N \operatorname{Softmax}\left(\operatorname{MLP}_1\left(\mathbf{V}_m^n\right)\right) \odot \mathbf{V}_m^n
\label{formula2}
\end{equation}
where $\mathbf{V}_m^n$ is feature node of $\widetilde{\mathcal{G}}_m$. MLP is a Multilayer Perceptron that assigns attention scores to $\mathbf{V}_m$. $\mathcal{H}_m$ is the feature representation of modality $m$.

\subsection{Bipartite Patient-Modality Graph}

Bipartite patient-modality graph can be denoted as $\mathcal{G}_P^M = \left\{\mathbf{V}_P, \mathbf{V}_M, \mathbf{A}\right\}$, the types of nodes consists of patient nodes $\mathcal{V}_P$ and modality nodes $\mathcal{V}_M$. $\mathcal{V}_P=\left\{u_1, \ldots, u_P\right\}$, $\mathcal{V}_M=\left\{v_1, \ldots, v_M\right\}$. $P$ and $M$ are the numbers of patients and modalities, respectively. In our study, there are 3 modalities, so $M$ = 3. The patient-modality relationship is presented as a patient-modality matrix $\mathbf{A}$. Specifically, if patient $p$ contains modality $m$, then the edge $e_{u_p v_m}$ is available. Otherwise, the edge is unavailable. For available edges, we initialize the edge embeddings using the extracted features of the corresponding available modalities. On the basis of the complete graph $\mathcal{G}_P^M$, an incomplete graph $\mathcal{G}_P^{M'}$ is obtained through random edge dropout to simulate different modality-missing scenarios. Then, we use a siamese Graph Neural Network (GNN) to encode the complete and incomplete graphs and integrate information from different modalities to obtain consolidated patient representations, as follows:

\begin{equation}
\begin{aligned}
\mathbf{Z}=\operatorname{GNN}(\mathcal{G}_P^M), \mathbf{Z}^{\prime}=\operatorname{GNN}(\mathcal{G}_P^{M'})
\label{formula3}
\end{aligned}
\end{equation}
where $\mathbf{Z}$ and $\mathbf{Z}^{\prime}$ are the representations of the same patient but with complete and incomplete modalities. To effectively predict survival risk, even when the modality is missing, the model should focus on features that are shared across modalities. Therefore, we propose a complete-incomplete alignment strategy ($\mathcal{L}_{\text {Cia }}$) to encourage the alignment of  $\mathbf{Z}$ and $\mathbf{Z}^{\prime}$ for uncovering modality-agnostic features. The specific implementation process is shown in Eq. \ref{formula5}.

\begin{equation}
\mathcal{L}_{Cia}\left(\mathbf{Z}, \mathbf{Z}^{\prime}\right)=-\sum_{p=1}^N \log \frac{e^{s\left(\mathbf{z}_p, \mathbf{z}_p^{\prime}\right) / \varphi}}{\sum_{q=1}^N e^{s\left(\mathbf{z}_p, \mathbf{z}_q^{\prime}\right) / \varphi}}
\label{formula5}
\end{equation}
where $\mathbf{z}_p$ and $\mathbf{z}_p^{\prime}$ are the features of same patient. $s(\cdot, \cdot)$ is the cosine similarity function, $\varphi$ is the temperature hyperparameter. It should be noted that this complete-incomplete alignment loss is not only applicable to uncensored data with accurate survival time but also to censored data.

After this, the representation of patient nodes $\mathbf{Z}$ is used to predict survival risk. After obtaining the survival risk, we use Cox loss ($\mathcal{L}_{\text {Cox }}$) as one of the optimization objectives. The definition of Cox loss is as follows:

\begin{equation}
\mathcal{L}_{\text {Cox }}=\sum_{i}^{B} \delta_{\mathrm{i}}\left(-\mathbf{O}(i)+\log \sum_{\mathrm{j}: \mathrm{t}_{\mathrm{j}} \geq \mathrm{t}_{\mathrm{i}}} \mathrm{e}^{\mathbf{O}(j)}\right)
\label{supp6}
\end{equation}
where $\mathbf{O}(i)$ is equal to $\theta x_{i}$ and $\mathbf{O}(j)$ is equal to $\theta x_{j}$, which represents the survival risks of the i-th sample and the j-th sample respectively.

\subsection{Event-Conditional Modelling of Censoring}

\noindent\textbf{Dynamic Momentum Accumulation Confidences.} To assess model confidence of each sample in survival prediction, we introduce a method based on the ranking difference between epochs. The difference serves as a stability criterion: larger differences indicate lower confidence, while smaller differences suggest higher confidence. Since confidence can fluctuate, especially in early epochs, we use the dynamic momentum accumulation confidence (DMAC) of each sample calculated at all previous epochs as the confidence of the model as follows:

\begin{equation}
\tau(t)=\lambda \tau(t-1)+(1-\lambda) p(t), \tau(0)=0
\label{formula10}
\end{equation}
where $p(t)$ is the reciprocal of the absolute value of the difference between the current epoch and the previous epoch, and $\lambda$ is the balance parameter that balances the confidence of the current epoch and all previous epochs.

\noindent\textbf{Selection and Update for Censored Data.} To enhance model performance, we select censored samples with high confidence for training. Specifically, all samples are first sorted by the predicted survival risk from CenSurv. Then, for each censored sample, with the goal of obtaining the best C-index within its K immediate range, the survival time and survival status of the sample are updated.  Finally, traverse and update the survival time and status of all selected censored samples. This process ensures the updated survival time exceeds the original censored time, and the status is updated to death from alive ($\delta$: 0 $\rightarrow$ 1). By iterating over all censored samples, we increase the proportion of censored samples in the training set, which ultimately improves the model performance.

\subsection{Optimization Objective}

In our study, the optimization objective includes Cox loss and Complete-incomplete alignment loss. The total loss of CenSurv is as follows:

\begin{equation}
\mathcal{L}_{All}=
\alpha \mathcal{L}_{Cox}+  \beta \mathcal{L}_{Cia}.
\label{formula11}
\end{equation}
where $\alpha$ and $\beta$ are the weights used to balance each loss. It is important to note that the optimization process consists of a preheating stage (0$\leq$epoch$ \textless$60) and an iterative update stage (60$\leq$epoch$ \textless$120). In the preheating stage, the status of censored and uncensored data is determined by the original clinical features. Once the model stabilizes, the process enters the iterative update stage, during which the survival events and survival statuses of the censored data are updated and treated as uncensored samples for inclusion in the training process.

\section{Experiments}
\subsection{Experimental Setting}

\noindent\textbf{Datasets.} We validate the effectiveness of CenSurv on 5 TCGA cancer datasets: Kidney Clear Cell Carcinoma (KIRC), Liver Hepatocellular Carcinoma (LIHC), Esophageal Carcinoma (ESCA), Lung Squamous Cell Carcinoma (LUSC), and Uterine Corpus Endometrial Carcinoma (UCEC).

\noindent\textbf{Implementation.} We use Adam as the optimizer, and its learning rate is 0.00003. After extensive experiments, the weights for the optimization objectives in this study are $\alpha$=5 and $\beta$=1. $\lambda$ in DMAC is set to 0.4. We use five-fold cross-validation to evaluate the performance of the proposed model on all 5 datasets, where the model achieving the best performance on the validation set in each fold is selected for testing. The results of all comparative experiments are reproduced based on the original settings and presented in terms of mean and standard deviation.

\noindent\textbf{Evaluation Metric.} The main evaluation metric used in our study is the Concordance index (C-index) \cite{harrell1982evaluating,steck2007ranking,shao2023hvtsurv}. We also calculate P-Value using the Logrank test to verify the model’s ability to distinguish between high and low risk groups.

\subsection{Results with Complete Modalities}

\noindent\textbf{Unimodal v.s. CenSurv.}  We implement Cox proportional hazards models in pathological data, genetic data, and clinical records. For pathological data, we compare the SOAT methods TranMIL~\cite{shao2021transmil} and HvtSurv~\cite{shao2023hvtsurv}. As shown in Table \ref{table1}, the mean C-index of CenSurv on 5 datasets is 5.7\% higher than the best comparative unimodal method. These results show that CenSurv can effectively fuse multi-modal features, thereby achieving more accurate survival prediction.

\begin{table*}[t!]
\footnotesize
        \caption{The C-index of multiple unimodal and multimodal methods on 5 datasets. The best results and the second-best results are highlighted in bold and in \underline{underline}.}
        \begin{center}
        \begin{tabular}{p{11pt}<{\centering}p{63pt}p{44.5pt}<{\centering}p{44.5pt}<{\centering}p{44.5pt}<{\centering}p{44.5pt}<{\centering}p{44.5pt}<{\centering}p{22pt}<{\centering}}
        \toprule
         &Method &KIRC & LIHC & ESCA & LUSC & UCEC & Mean\\
          \midrule 
     \multirow{5}*{\rotatebox{90}{Unimodal}}
     &Cox$\operatorname{(c)}$\cite{katzman2018deepsurv}        &0.687 $_{ 0.012}$ & 0.513 $_{ 0.031}$ & 0.585 $_{ 0.030}$ & 0.559 $_{ 0.009}$ & 0.675 $_{ 0.017}$ & 0.605\\
    &Cox$\operatorname{(g)}$\cite{katzman2018deepsurv}    &0.665 $_{ 0.008}$ & 0.649 $_{ 0.013}$ & 0.533 $_{ 0.020}$ & 0.519 $_{ 0.017}$ & 0.686 $_{ 0.027}$ & 0.610\\

&Cox$\operatorname{(p)}$\cite{katzman2018deepsurv}   &0.647 $_{ 0.009}$ & 0.635 $_{ 0.012}$ & 0.620 $_{ 0.017}$ & 0.551 $_{ 0.007}$ & 0.662 $_{ 0.015}$& 0.603 \\

&TranMIL$\operatorname{(p)}$\cite{shao2021transmil}  &0.674 $_{ 0.012}$  &0.657 $_{ 0.016}$  &0.568 $_{ 0.016}$  &0.551 $_{ 0.009}$  &0.686 $_{ 0.020}$& 0.627 \\

&HvtSurv$ \operatorname{(p)}$\cite{shao2023hvtsurv}   &0.698 $_{ 0.009}$  &0.664 $_{ 0.015}$  &0.622 $_{ 0.014}$  &0.564 $_{ 0.013}$  &0.709 $_{ 0.019}$& 0.651 \\
 
            \midrule 
\multirow{9}*{\rotatebox{90}{Multimodal}}
 &GSCNN\cite{mobadersany2018predicting}  & 0.702 $_{0.011}$  & 0.651 $_{0.009}$ & 0.554 $_{0.012}$ & 0.531 $_{0.010}$ & 0.723 $_{0.012}$& 0.632\\
&MultiSurv\cite{vale2021long}    & 0.662 $_{0.015}$ & 0.634 $_{0.009}$ & 0.573 $_{0.019}$ & 0.557 $_{0.016}$ & 0.693 $_{0.025}$ & 0.623\\

&OuterP\cite{chen2020pathomic}    & 0.686 $_{0.013}$ & 0.664 $_{0.021}$ & 0.610 $_{0.019}$ & 0.559 $_{0.012}$ & 0.681 $_{0.026}$ & 0.640\\

&MetricL\cite{cheerla2019deep}     & 0.730 $_{0.022}$ & 0.659 $_{0.019}$ & 0.615 $_{0.022}$ & \underline{0.576} $_{0.015}$ & 0.696 $_{0.010}$ & 0.655 \\

&MCAT\cite{chen2021multimodal}    & 0.681 $_{0.011}$ & 0.677 $_{0.012}$ & 0.576 $_{0.016}$ & 0.573 $_{0.014}$ & 0.691 $_{0.026}$ & 0.640\\

&HGCN \cite{hou2023hybrid}   &\underline{0.742} $_{0.008}$ & 0.692 $_{0.011}$ & 0.627 $_{0.017}$ & 0.569 $_{0.014}$ & 0.742 $_{0.015}$ & 0.674 \\ 

&SurvMamba\cite{chen2024survmamba}   & 0.727 $_{0.009}$ & 0.690 $_{0.013}$ & \underline{0.633} $_{0.015}$ & 0.566 $_{0.016}$ & {0.746} $_{0.016}$ & {0.672} \\ 

&SurvPath\cite{jaume2023modeling}    & 0.739 $_{0.011}$ & \underline{0.699} $_{0.010}$ & {0.631} $_{0.013}$ & 0.570 $_{0.013}$ & \underline{0.746} $_{0.013}$ & \underline{0.677} \\ 

&\textbf{CenSurv}   &\textbf{0.751} $_{0.009}$  &\textbf{0.708} $_{0.010}$  &\textbf{0.691} $_{0.015}$  &\textbf{0.607} $_{0.015}$  &\textbf{0.781} $_{0.013}$ & \textbf{0.708} \\
                \bottomrule
        \end{tabular}
        \end{center}
        \label{table1}
\end{table*}

\noindent\textbf{Multimodal v.s. CenSurv.} We compare CenSurv with GSCNN~\cite{mobadersany2018predicting}, MultiSurv~\cite{vale2021long}, OuterP~\cite{chen2020pathomic}, MetricL~\cite{cheerla2019deep}, MCAT~\cite{chen2021multimodal}, HGCN~\cite{hou2023hybrid}, SurvMamba~\cite{chen2024survmamba} and SurvPath~\cite{jaume2023modeling}. Compared with SOAT multimodal methods, CenSurv can also achieve the best performance. Specifically, the best-performing multimodal comparative method achieves a mean C-index of 0.677 on 5 datasets, while Censurv achieves a mean C-index of 0.708 on the same 5 datasets, exceeding 3.1\%. In addition, the P-Values of CenSurv for distinguishing high and low risks on the 5 datasets are 6.92$\times$10$^{-15}$, 3.50$\times$10$^{-6}$, 5.31$\times$10$^{-4}$, 1.57$\times$10$^{-3}$ and 7.41$\times$10$^{-8}$, respectively, which are better than the other comparison methods. These results show that CenSurv outperforms other multimodal methods. In addition, CenSurv contains 2,308.91K parameters and requires 2.26G FLOPs. While not the most lightweight in terms of model complexity, it achieves a good balance between computational cost and predictive performance.

\begin{figure}[t!]
\centerline{\includegraphics[width=0.7\textwidth]{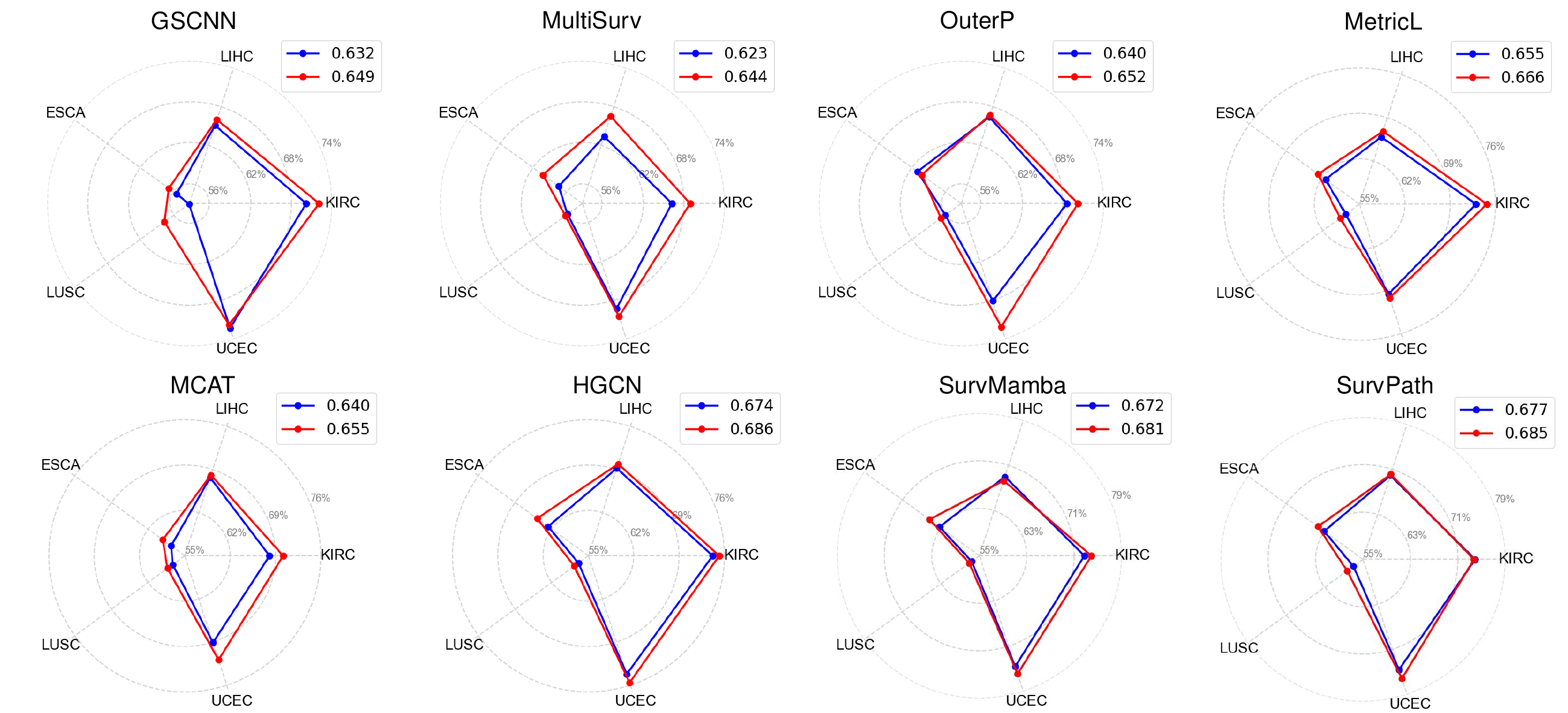}}
\caption{The C-index of compared multimodal methods with and without ECMC. Red represents the methods with ECMC, and blue represents the methods without ECMC.}
\label{ecmc}
\end{figure}

\noindent\textbf{Robustness Analysis of ECMC.} To verify the robustness of the plug-and-play module ECMC, we conduct experimental comparisons with and without ECMC in 8 multimodal survival prediction methods. As seen in Fig. \ref{ecmc}, as a plug-and-play module, after adding the ECMC module, the mean C-index across 5 datasets improved for all methods. For example, compared with the traditional MultiSurv, MultiSurv with the ECMC module improves the mean C-index by 2.1\% on 5 datasets. All analyses show the superiority and robustness of ECMC.

\subsection{Results with Incomplete Modalities}

\begin{figure}[t!]
\centerline{\includegraphics[width=0.7\textwidth]{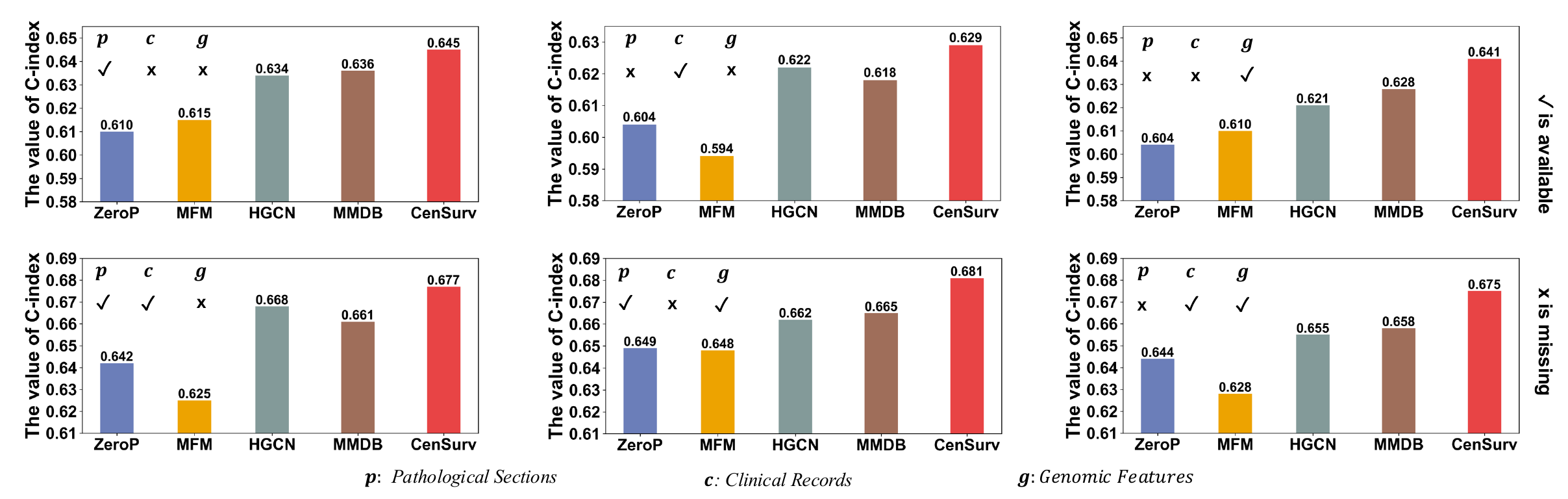}}
\caption{The mean C-index of methods under various modality-missing scenarios.}
\label{incomplete}
\end{figure}

We compare CenSurv with existing methods for handling missing modalities, including ZeroP~\cite{shen2020memor}, MFM~\cite{tsai2018learning}, and HGCN~\cite{hou2023hybrid} and MMDB~\cite{qu2024multi}. As shown in Fig. \ref{incomplete},  CenSurv shows excellent performance in different missing-modality scenarios, and its mean C-index on 5 datasets is ahead of other methods.

\subsection{Ablation Studies}
To validate the effectiveness of each module in our proposed method, we sequentially remove each component (ECMC, BPMG, DMAC) while keeping the other components unchanged. As shown in Table \ref{table4}, as a plug-and-play module, adding the ECMC can improve the performance of CenSurv on all 5 datasets. We designed the BPMG module, which not only enables the model to adapt to various modality-missing scenarios but also enhances the performance across most datasets. Using the DMAC module to select samples with higher comprehensive confidence can avoid adding noisy samples and thus improve model performance. The above analyses also show the effectiveness of each module.

\begin{table}[t!]
\footnotesize
        \caption{The ablation study on 5 datasets, the best results are highlighted in bold.}
        \begin{center}
        \begin{tabular}{p{63pt}|p{45pt}<{\centering}p{45pt}<{\centering}p{45pt}<{\centering}p{45pt}<{\centering}p{45pt}<{\centering}p{26.4pt}<{\centering} }
        \toprule
          & KIRC &  LIHC &  ESCA &  LUSC  & UCEC & Mean\\
          \midrule 
w/o ECMC    & 0.745 $_{0.008}$  & 0.691 $_{0.011}$ & 0.643 $_{0.014}$  & 0.567 $_{0.013}$  & 0.747 $_{0.010}$ & 0.678\\
w/o BPMG    & 0.747 $_{0.011}$  & 0.699 $_{0.009}$ & \textbf{0.694} $_{0.012}$  & 0.585 $_{0.013}$  & 0.765 $_{0.014}$ & 0.698 \\
w/o DMAC    & 0.733 $_{0.010}$  & 0.662 $_{0.015}$ & 0.637 $_{0.013}$  & 0.557 $_{0.014}$  & 0.745 $_{0.015}$  &0.667\\
\textbf{CenSurv} &\textbf{0.751} $_{0.009}$  &\textbf{0.708} $_{0.010}$  &0.691 $_{0.015}$  &\textbf{0.607} $_{0.015}$  &\textbf{0.781} $_{0.013}$ & \textbf{0.708} \\
\bottomrule
        \end{tabular}
        \end{center}
        \label{table4}
 \end{table}

We conduct a quality analysis of updated survival times with censored data by randomly pruning the true survival time between 0 and the true value to simulate censored data. The initial MAE for both uncensored and censored data is 14.5 months, decreasing to 5.7 months after applying ECMC, demonstrating its ability to provide more accurate survival times. In addition, some hyperparameters, such as $\alpha$, $\beta$, and $\lambda$, are also determined through additional experiments.

\section{Conclusion}

In our study, we propose a bipartite patient-modality graph learning with event-conditional censoring modeling for survival prediction. Overall, CenSurv offers a novel approach for cancer survival prediction. In future work, we will enhance interpretability across multiple modalities, investigate model complexity, and incorporate additional evaluation metrics.

\noindent\textbf{Acknowledgements.} This work was supported in part by Xinjiang Uygur Autonomous Region Key R\&D program (No. 2024B03039-3), the central government guides local funds for scientific and technological development of China (No. ZYYD2025QY25), the Science and Technology Major Project of Changsha (No. kh2502004), Scientific Research Fund of Hunan Provincial Education Department (No. 23A0020), and the Fundamental Research Funds for the Central Universities of Central South University (No. 2024ZZTS0106). This work was carried out in part using computing resources at the High-Performance Computing Center of Central South University.

\noindent\textbf{Disclosure of Interests.} The authors have no competing interests to declare that are relevant to the content of this article.

\bibliographystyle{splncs04.bst}
\bibliography{Paper-2356.bib}

\begin{thebibliography}{10}
\providecommand{\url}[1]{\texttt{#1}}
\providecommand{\urlprefix}{URL }
\providecommand{\doi}[1]{https://doi.org/#1}

\bibitem{cheerla2019deep}
Cheerla, A., Gevaert, O.: Deep learning with multimodal representation for pancancer prognosis prediction. Bioinformatics  \textbf{35}(14),  i446--i454 (2019)

\bibitem{chen2020pathomic}
Chen, R.J., Lu, M.Y., Wang, J., Williamson, D.F., Rodig, S.J., Lindeman, N.I., Mahmood, F.: Pathomic fusion: an integrated framework for fusing histopathology and genomic features for cancer diagnosis and prognosis. IEEE Transactions on Medical Imaging  \textbf{41}(4),  757--770 (2020)

\bibitem{chen2021multimodal}
Chen, R.J., Lu, M.Y., Weng, W.H., Chen, T.Y., Williamson, D.F., Manz, T., Shady, M., Mahmood, F.: Multimodal co-attention transformer for survival prediction in gigapixel whole slide images. In: Proceedings of the IEEE/CVF International Conference on Computer Vision. pp. 4015--4025 (2021)

\bibitem{chen2024survmamba}
Chen, Y., Xie, J., Lin, Y., Song, Y., Yang, W., Yu, R.: Survmamba: State space model with multi-grained multi-modal interaction for survival prediction. arXiv preprint arXiv:2404.08027  (2024)

\bibitem{elforaici2025semi}
Elforaici, M.E.A., Montagnon, E., Romero, F.P., Le, W.T., Azzi, F., Trudel, D., Nguyen, B., Turcotte, S., Tang, A., Kadoury, S.: Semi-supervised vit knowledge distillation network with style transfer normalization for colorectal liver metastases survival prediction. Medical Image Analysis  \textbf{99},  103346 (2025)

\bibitem{hamilton2017inductive}
Hamilton, W., Ying, Z., Leskovec, J.: Inductive representation learning on large graphs. Advances in neural information processing systems  \textbf{30} (2017)

\bibitem{harrell1982evaluating}
Harrell, F.E., Califf, R.M., Pryor, D.B., Lee, K.L., Rosati, R.A.: Evaluating the yield of medical tests. Jama  \textbf{247}(18),  2543--2546 (1982)

\bibitem{hou2023hybrid}
Hou, W., Lin, C., Yu, L., Qin, J., Yu, R., Wang, L.: Hybrid graph convolutional network with online masked autoencoder for robust multimodal cancer survival prediction. IEEE Transactions on Medical Imaging  \textbf{42}(8),  2462--2473 (2023)

\bibitem{jaume2023modeling}
Jaume, G., Vaidya, A., Chen, R., Williamson, D., Liang, P., Mahmood, F.: Modeling dense multimodal interactions between biological pathways and histology for survival prediction. Proceedings of the IEEE/CVF Conference on Computer Vision and Pattern Recognition (CVPR)  (2024)

\bibitem{katzman2018deepsurv}
Katzman, J.L., Shaham, U., Cloninger, A., Bates, J., Jiang, T., Kluger, Y.: Deepsurv: personalized treatment recommender system using a cox proportional hazards deep neural network. BMC medical research methodology  \textbf{18},  1--12 (2018)

\bibitem{lee2024machine}
Lee, K.H., Choi, G.H., Yun, J., Choi, J., Goh, M.J., Sinn, D.H., Jin, Y.J., Kim, M.A., Yu, S.J., Jang, S., et~al.: Machine learning-based clinical decision support system for treatment recommendation and overall survival prediction of hepatocellular carcinoma: a multi-center study. npj Digital Medicine  \textbf{7}(1), ~2 (2024)

\bibitem{liu2024agnostic}
Liu, H., Shi, Y., Xu, Y., Li, A., Wang, M.: Agnostic-specific modality learning for cancer survival prediction from multiple data. IEEE Journal of Biomedical and Health Informatics  (2024)

\bibitem{mobadersany2018predicting}
Mobadersany, P., Yousefi, S., Amgad, M., Gutman, D.A., Barnholtz-Sloan, J.S., Vel{\'a}zquez~Vega, J.E., Brat, D.J., Cooper, L.A.: Predicting cancer outcomes from histology and genomics using convolutional networks. Proceedings of the National Academy of Sciences  \textbf{115}(13),  E2970--E2979 (2018)

\bibitem{psutka2022clinical}
Psutka, S.P., Gulati, R., Jewett, M.A., Fadaak, K., Finelli, A., Legere, L., Morgan, T.M., Pierorazio, P.M., Allaf, M.E., Herrin, J., et~al.: A clinical decision aid to support personalized treatment selection for patients with clinical t1 renal masses: Results from a multi-institutional competing-risks analysis. European urology  \textbf{81}(6),  576--585 (2022)

\bibitem{qu2024multi}
Qu, L., Huang, D., Zhang, S., Wang, X.: Multi-modal data binding for survival analysis modeling with incomplete data and annotations. In: International Conference on Medical Image Computing and Computer-Assisted Intervention. pp. 501--510. Springer (2024)

\bibitem{riasatian2021fine}
Riasatian, A., Babaie, M., Maleki, D., Kalra, S., Valipour, M., Hemati, S., Zaveri, M., Safarpoor, A., Shafiei, S., Afshari, M., et~al.: Fine-tuning and training of densenet for histopathology image representation using tcga diagnostic slides. Medical image analysis  \textbf{70},  102032 (2021)

\bibitem{shahin2024centime}
Shahin, A.H., Zhao, A., Whitehead, A.C., Alexander, D.C., Jacob, J., Barber, D.: Centime: Event-conditional modelling of censoring in survival analysis. Medical Image Analysis  \textbf{91},  103016 (2024)

\bibitem{shao2023fam3l}
Shao, W., Liu, J., Zuo, Y., Qi, S., Hong, H., Sheng, J., Zhu, Q., Zhang, D.: Fam3l: Feature-aware multi-modal metric learning for integrative survival analysis of human cancers. IEEE Transactions on Medical Imaging  \textbf{42}(9),  2552--2565 (2023)

\bibitem{shao2021transmil}
Shao, Z., Bian, H., Chen, Y., Wang, Y., Zhang, J., Ji, X., et~al.: Transmil: Transformer based correlated multiple instance learning for whole slide image classification. Advances in neural information processing systems  \textbf{34},  2136--2147 (2021)

\bibitem{shao2023hvtsurv}
Shao, Z., Chen, Y., Bian, H., Zhang, J., Liu, G., Zhang, Y.: Hvtsurv: Hierarchical vision transformer for patient-level survival prediction from whole slide image. In: Proceedings of the AAAI Conference on Artificial Intelligence. vol.~37, pp. 2209--2217 (2023)

\bibitem{shen2020memor}
Shen, G., Wang, X., Duan, X., Li, H., Zhu, W.: Memor: A dataset for multimodal emotion reasoning in videos. In: Proceedings of the 28th ACM international conference on multimedia. pp. 493--502 (2020)

\bibitem{steck2007ranking}
Steck, H., Krishnapuram, B., Dehing-Oberije, C., Lambin, P., Raykar, V.C.: On ranking in survival analysis: Bounds on the concordance index. Advances in neural information processing systems  \textbf{20} (2007)

\bibitem{subramanian2005gene}
Subramanian, A., Tamayo, P., Mootha, V.K., Mukherjee, S., Ebert, B.L., Gillette, M.A., Paulovich, A., Pomeroy, S.L., Golub, T.R., Lander, E.S., et~al.: Gene set enrichment analysis: a knowledge-based approach for interpreting genome-wide expression profiles. Proceedings of the National Academy of Sciences  \textbf{102}(43),  15545--15550 (2005)

\bibitem{sung2021global}
Sung, H., Ferlay, J., Siegel, R.L., Laversanne, M., Soerjomataram, I., Jemal, A., Bray, F.: Global cancer statistics 2020: Globocan estimates of incidence and mortality worldwide for 36 cancers in 185 countries. CA: a cancer journal for clinicians  \textbf{71}(3),  209--249 (2021)

\bibitem{tsai2018learning}
Tsai, Y.H.H., Liang, P.P., Zadeh, A., Morency, L.P., Salakhutdinov, R.: Learning factorized multimodal representations. arXiv preprint arXiv:1806.06176  (2018)

\bibitem{vale2021long}
Vale-Silva, L.A., Rohr, K.: Long-term cancer survival prediction using multimodal deep learning. Scientific Reports  \textbf{11}(1),  13505 (2021)

\end{thebibliography}

\end{document}